\documentclass[12pt,a4paper]{cibb}

\usepackage{subfigure,graphicx}
\usepackage{wrapfig}
\makeatletter
  
  \renewcommand\p@subfigure{}       
\makeatother

\usepackage{amsmath,amsfonts,latexsym,amssymb,euscript,xr}
\usepackage{booktabs}
\usepackage[nodayofweek]{datetime}
\usepackage{hyperref}

\usepackage[table]{xcolor}
\usepackage{color,colortbl,tabularx}

\usepackage[english]{babel}
\usepackage[protrusion=true,expansion=true]{microtype}
\usepackage{amsmath,amsfonts,amsthm}
\usepackage{pifont}

\definecolor{LightBlue}{rgb}{0.88,0.9,0.9}

\usepackage{todonotes}

\title{A Machine Learning Pipeline for Multiple Sclerosis Biomarker Discovery: Comparing explainable AI and Traditional Statistical Approaches}
\author{\large Samuele Punzo$^{1,*}$, Silvia Giulia Galfrè$^{1,*}$, Francesco Massafra$^{1}$, Alessandro Maglione$^{2}$, Corrado Priami$^{1}$, Alina S\^irbu$^{3}$}
\address{\footnotesize $\ $\\$^1$ Department of Computer Science , University of Pisa,
Italy. \\
$^2$ Department of Clinical and Biological Sciences, University of Torino, Italy --
Department of Computer Sciences, University of Torino, Italy\\
$^3$ Department of Computer Science and Engineering, University of Bologna,
Italy.\\
\bigskip
$^*$corresponding author
}

\abstract{\small Machine Learning, Explainable AI, Differential expression, Multiple Sclerosis \normalsize
\\[17pt]
{\bf Abstract.} 
We present a machine learning pipeline for biomarker discovery in Multiple Sclerosis (MS), integrating eight publicly available microarray datasets from Peripheral Blood Mononuclear Cells (PBMC). After robust preprocessing we trained an XGBoost classifier optimized via Bayesian search. SHapley Additive exPlanations (SHAP) were used to identify key features for model prediction, indicating thus possible biomarkers. These were compared with genes identified through classical Differential Expression Analysis (DEA). Our comparison revealed both overlapping and unique biomarkers between SHAP and DEA, suggesting complementary strengths. Enrichment analysis confirmed the biological relevance of SHAP-selected genes, linking them to pathways such as sphingolipid signaling, Th1/Th2/Th17 cell differentiation, and Epstein-Barr virus infection—all known to be associated with MS. This study highlights the value of combining explainable AI (xAI) with traditional statistical methods to gain deeper insights into disease mechanism.
}

\begin{document}
\thispagestyle{myheadings}
\pagestyle{myheadings}

\section{\bf Introduction}
\label{sec:Introduction}
Multiple Sclerosis (MS) is a complex autoimmune disorder lacking reliable blood‐based biomarkers. Although traditional differential expression analysis of microarray data has identified candidate genes, it may overlook multivariate patterns critical to the disease. Here, we present an end-to-end pipeline that integrates eight PBMC microarray datasets, applies robust normalization, ComBat batch correction and trains a Bayesian-optimized XGBoost classifier. We use SHAP to interpret model decisions and directly compare these findings with canonical DEA. This approach not only confirms known MS‐related genes but also uncovers novel candidates, demonstrating how explainable AI enhances classical statistical methods.


\section{\bf Data and Methods}
\label{sec:DATA-AND-METHODS}
For this study, 8 publicly available PBMC microarray datasets (GSE41848, GSE41849, GSE146383, GSE13732, GSE136411, GSE17048, GSE41890, GSE21942)
have been selected from the NCBI Gene Expression Omnibus database to form a single dataset of interest, maintaining only the common genes. Each dataset includes samples from both healthy individuals (Control) and individuals diagnosed with multiple sclerosis (MS). All samples were derived from adults, and many individuals had multiple follow-up. While the MS samples cover various disease stages, they were all labelled simply as MS since differentiating them was beyond the scope of this study. Importantly, none of the MS individuals had received any form of therapy.

\begin{wrapfigure}{L}{0.5\linewidth}
  \includegraphics[width=0.5\textwidth]{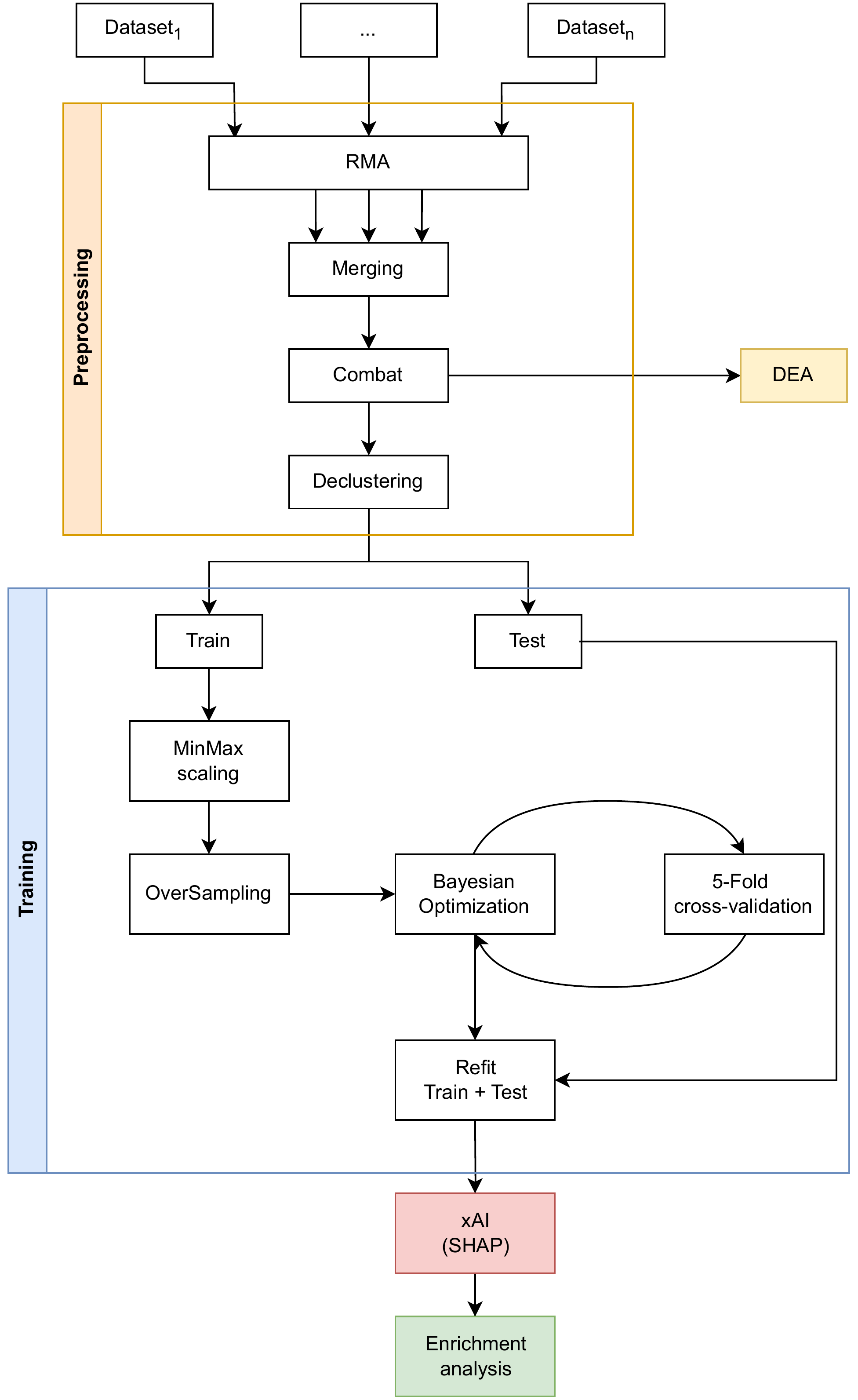}
    \caption{Representation of the analysis pipeline.}
    \label{fig:Pipeline}
\end{wrapfigure}

The implemented pipeline is shown in \autoref{fig:Pipeline} and can be split into: Preprocessing, Training and Explainability.
The analyses were performed in R and Python, both data and analyses are publicly available on \href{https://github.com/seriph78/ML_for_MS.git}{GitHub}.

\newpage
\subsection{Preprocessing}
We harmonized the datasets via RMA normalization (background correction, quantile normalization, probe summarization, $log_2$ transformation). ComBat’s empirical Bayes approach was then applied to correct batch effects, and all gene features were scaled to $[0, 1]$ using MinMax normalization.  To evaluate preprocessing quality, both qualitative and quantitative techniques were employed, including: DEA between datasets under identical conditions, visual assessment of sample distribution via Principal Component Analysis (PCA), analysis of multimodal distributions, and Mixture Score quantification~\cite{MixtureScore}. 
To reduce feature redundancy and simplify interpretation, we collapsed clusters of highly correlated genes (Pearson’s $r > 0.9$) into a single representative—the gene with highest variance in each cluster. This declustering minimizes collinearity and facilitates downstream ML analyses.

\subsection{Training}
 We split the declustered data into training and test sets ($75\%/25\% $), keeping all samples from each subject together, considering the different follow-ups while also stratifying the split so that each partition contain the same proportion of samples from each dataset. To address class imbalance in the training data, we applied SMOTE~\cite{SMOTE} (via imbalanced-learn) to synthetically balance minority and majority classes. An XGBoost classifier was then trained with Bayesian hyperparameter optimization and 5-fold cross-validation. This strategy efficiently explores the hyperparameter space and yields a robust performance estimate. Due to class imbalance, F1-score was selected as the metric for training, which allowed us to balance both precision and recall, ensuring relevant predictions for the minority class (Control). Finally, the best model was evaluated on the hold-out test set (see \autoref{ssec:trainingResults}) and after being retrained on the wholes dataset (train + test) it was subjected to the explainability analysis.


\subsection{Explainable AI \& Differential Expression Analysis}


We computed SHAP values using TreeExplainer~\cite{TreeExplainer} to rank features by their contribution to XGBoost predictions. Genes with non-zero importance were selected, and we included back the genes in their respective clusters, 
assigning the same importance score as the representative gene. In parallel, we performed a Differential Expression Analysis (Wilcoxon rank-sum test, adjusted by Benjamini–Hochberg $FDR < 0.05$) to identify genes significantly altered between MS and Control samples. 



\begin{figure}[h]
    \centering
    \subfigure{\includegraphics[width=0.4\textwidth]{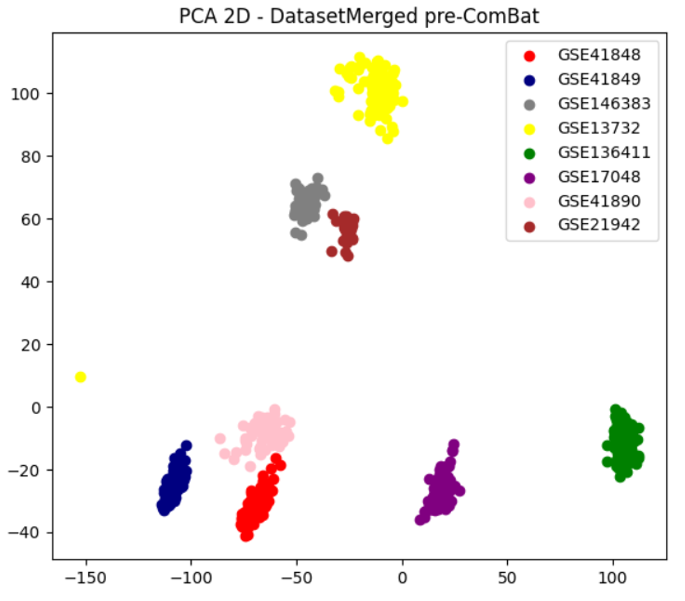}}
    \subfigure{\includegraphics[width=0.4\textwidth]{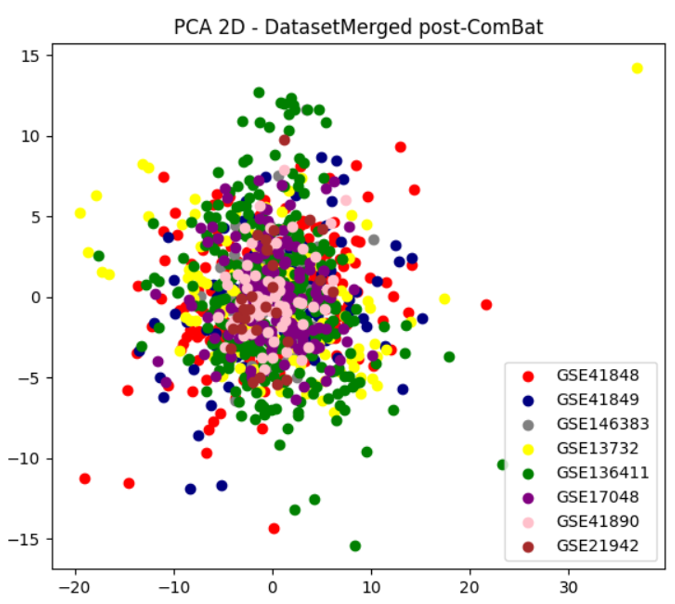}}
    \caption{PCA pre and post ComBat normalisation. Different colours correspond to samples from different datasets.}
    \label{fig:PCA_prepostCombat}
\end{figure}

\section{Results}

\subsection{Preprocessing}
After normalization and merging, the dataset is formed of $363$ Control and $676$ MS samples for a total of $1038$ samples with $6673$ features;
the inclusion of GSE136411 dataset significantly reduced the number of common genes ($6673$ vs $11492$) but greatly increased the number of samples ($729$ vs $1038$). This occurs because GSE136411 was obtained with a microarray whose manufacturer differs from the others (Illumina vs Affymetrix).
\par After merging the datasets, we applied ComBat normalization. In Figure \ref{fig:PCA_prepostCombat} we show, with the help of PCA, how the shape of the sample distribution has changed with the removal of batch effect compared to the original data. Other than this visual analysis, a DEA was computed between the single datasets previously divided by condition, resulting in no genes reaching statistical significance to be differentially expressed between datasets in the two groups (Control and MS). We also employed the Mixture Score, a quantitative measure to evaluate the quality of batch effect removal, resulting in a mean score (calculated over the two conditions) of 0.29, where 0.50 is the ideal~\cite{MixtureScore}. Additionally, an analysis of multimodal expression was made, showing the following genes as potentially multimodal: 
PRKY, HLA-DRB1, TMSB4Y, RPS21, HLA-DRB5, ZNF549, RPS4Y1, DDX3Y, UTY, EIF1AY, TXLNGY.
We can observe that some of these genes are Y-linked, with their multimodality being caused by gender-dependent expression patterns. Furthermore, genes such as HLA-DRB1, HLA-DRB5 are associated with MS risk, and involved in immune responses\cite{baranziniGeneticsMultipleSclerosis2017}. For the others, after careful examination, no sign of evident difference between datasets has been observed, so all of them have been maintained. The gene correlation analysis identified only one significant cluster, containing Y-linked genes: DDX3Y, EIF1AY, RPS4Y1, TXLNGY, UTY, PRKY, EIF1AY. We selected as representative the EIF1AY gene (highest variance in the cluster).

\begin{table}[h]
\footnotesize
    \centering
    \caption{Hyperparameter search space.}
    \begin{tabular}{c|c||c|c}
    \textbf{Hyperparameter} & \textbf{Range} &  \textbf{Hyperparameter} & \textbf{Range} \\
    \hline
        \textit{n\_estimators} & $50$ to $600$ &    \textit{reg\_alpha} & $0.0001$ to $100$ \\
    \hline
        \textit{max\_depth} & $2$ to $15$ & \textit{reg\_lambda} & $0.0001$ to $100$\\
    \hline
        \textit{learning\_rate} & $0.001$ to $1$ & \textit{min\_child\_weight} & $1$ to $10$ \\
    \hline
        \textit{gamma} & $0.0001$ to $100$ & \textit{scale\_pos\_weight} & $1$, $400/510$ \\
    \hline
    \end{tabular}
    \label{tab:hyperparameters}
\end{table}

\clearpage
\subsection{Training and model selection}
\label{ssec:trainingResults}

\begin{wrapfigure}{L}{0.55\linewidth}
    \centering
    \includegraphics[width=0.5\textwidth]{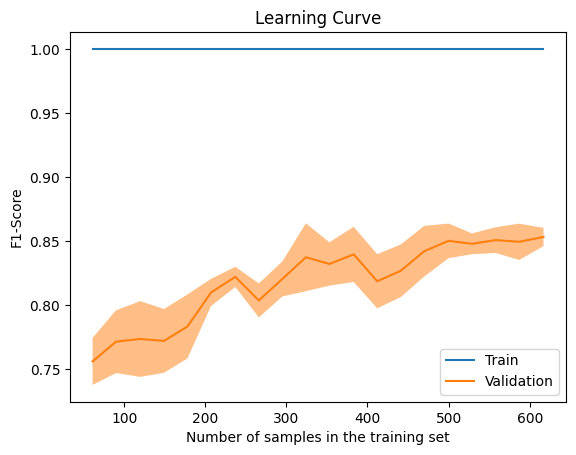}
    \caption{Learning curve of XGB classifier.}
     \label{fig:XGB_classifier}
\end{wrapfigure}

For the training phase, the declustered dataset was split into training ($510$ Control and $262$ MS samples) and test ($166$ Control and $100$ MS samples) sets. We oversampled the training set using a sampling ratio of $400$ controls to $510$ MS samples and $k=5$ nearest neighbors, which generated $138$ synthetic samples for a more balanced representation. We then trained the model with a Bayesian search with $600$ iterations over the hyperparameter search space detailed in \autoref{tab:hyperparameters}. The prediction performance of the best model are shown in~\autoref{tab:XGBresults}. We note that, in general, prediction performance is good. However, some difference between train and test performance is visible, indicating some overfitting. This was also present for validation data, with a mean validation F1-score of 0.78 over the 5 cross-validation folds. Hence, we believe it is due to the relatively low number of samples available. Figure~\ref{fig:XGB_classifier} displays the model’s learning curves, clearly indicating that additional data would likely improve validation performance, as evidenced by the upward trend in the validation curve with increasing sample size. The ROC curve yields an AUC of 0.86 on test data. All results demonstrate good sensitivity, specificity, and effective class separation, sufficient for the main aim of our analysis, which is the identification of possible biomarkers.

\begin{table}[t]
\footnotesize
 \centering
    \caption{Performance of the best Gradient Boosting model. Where applicable, values shown are macro averages over the two classes.}
    \begin{tabular}{c||c||c|c|c|c}
    \textbf{Train F1-Score} & \textbf{Validation F1-Score}&\textbf{Test F1-score} & \textbf{ Test Accuracy} & \textbf{Test Precision} & \textbf{Test Recall}\\
    \hline
     1.0 & 0.78 & 0.75 & 0.78 & 0.79 & 0.74\\
    \hline
    \end{tabular}
    \label{tab:XGBresults}
\end{table}

\begin{wrapfigure}{L}{0.4\linewidth}
    \centering
    \includegraphics[width=0.35\textwidth]{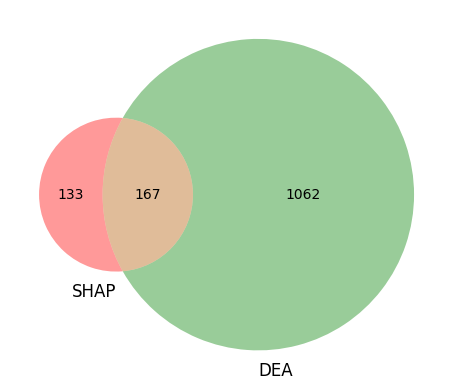}
    \caption{Venn diagram of SHAP features and DE features.}
    \label{fig:DEA_vs_Shap}
\end{wrapfigure}

\subsection{DEA vs SHAP}

The  genes identified by SHAP and DEA included an
overlap of 167 genes (Figure~\ref{fig:DEA_vs_Shap}). 133 genes were unique to SHAP, while DEA  was much less specific and identified over 1000 additional genes.
Both methods identified known MS-associated genes HLA-DRB1 and HLA-DRB5, which ranked among the top five in SHAP importance. However, other genes with known association,  GABPA, GFI1 (SHAP detects GFI1B, a strictly related gene), and MBP were found only through DEA, while SHAP uniquely highlighted EGR1, IL1B, and IL2RA. IL1B is a mediator of inflammation. Its blockade attenuated the disease in a pre-clinical model of MS\cite{linNewInsightsRole2017} while IL2RA is another well-known susceptibility gene, that was also used as drug target in MS\cite{weberIL2RAIL7RAGenes2008, baranziniGeneticsMultipleSclerosis2017}. 
These results demonstrate that xAI methods like SHAP can confirm known findings while also uncovering complementary and potentially novel insights compared to canonical approaches.




\subsection{Biological interpretation}
\begin{figure}[!h]
    \centering
    \subfigure[Zoomed STRING network of SHAP‐identified genes, with RUNX1 transcriptional regulation in pink, pathways from Devanand et al.~\cite{devanandSignalingMechanismsInvolved2023} in blue/violet, sphingolipid signaling in green, and Epstein–Barr virus infection in red.\label{fig:stringEnrichmentZoom}]{\includegraphics[width=0.57\linewidth]{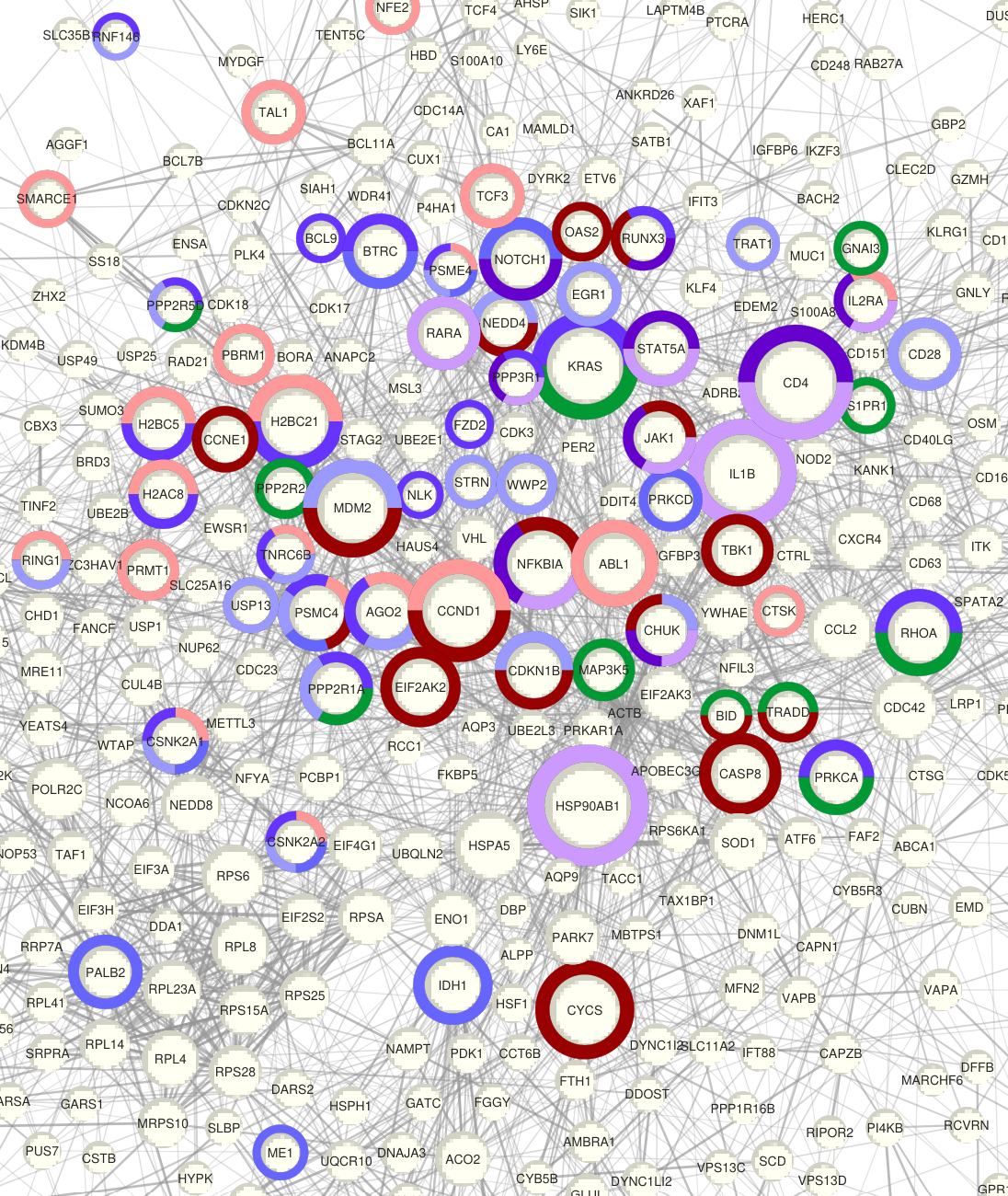}}
    \hfill
    \subfigure[SHAP value distributions for ABCA1, NDUFS5 and EIF2S2 (red = MS, blue = Control), highlighting enrichment of ABCA1 and NDUFS5 and depletion of EIF2S2 in MS samples.\label{fig:NewGenes}]{\includegraphics[width=0.35\linewidth]{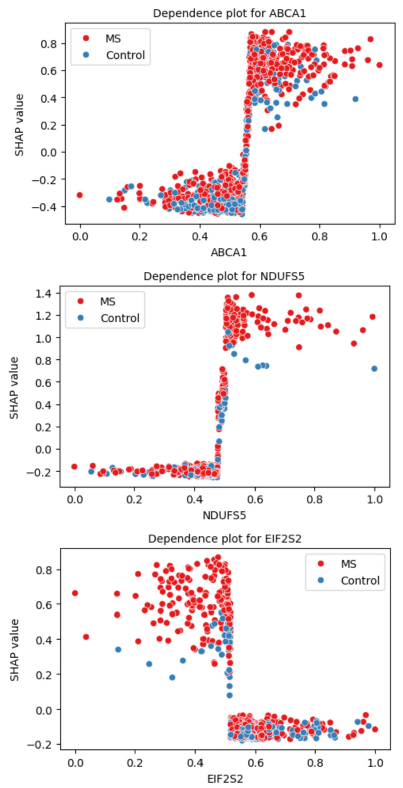}}
    \caption{Biological results}
    \label{fig:BioResults}
    
\end{figure}

We analyzed the SHAP-prioritized genes using Cytoscape and STRING to construct a gene interaction network. Most genes formed a connected network, with only 89 out of 698 remaining isolated (Figure~\ref{fig:BioResults}\ref{fig:stringEnrichmentZoom}). Enrichment analysis revealed significant associations (FDR $< 0.05$) with MS-relevant pathways, including sphingolipid signaling, which can be linked to inflammation~\cite{maceykaSphingolipidMetabolitesInflammatory2014}, Th1/Th2/Th17 cell differentiation, and Epstein-Barr virus infection—each well established in the MS literature~\cite{ascherioEpsteinBarrVirus2010}. Additional enriched pathways involved KEAP1-NFE2L2 signaling, WNT and AKT signaling indicated in the work of Devanand et al.~\cite{devanandSignalingMechanismsInvolved2023}, and transcriptional regulation by RUNX1, a gene with potential diagnostic and prognostic value for MS~\cite{haridyDiagnosticPrognosticValue2023}.

Among the top-ranked SHAP genes were: ABCA1, key to phospholipid transfer and HDL formation; NDUFS5, a non-catalytic subunit of mitochondrial NADH dehydrogenase;
EIF2S2, part of the eIF2 complex involved in early-stage protein synthesis; EIF2S2, may act as a functional link between lipid transport and mitochondrial stress response pathways, suggesting a broader regulatory role in MS pathogenesis. Figure~\ref{fig:BioResults}\ref{fig:NewGenes} shows SHAP values for these genes, per patient. We note how the first two genes appear to be promoters of the disease, with higher expression values corresponding to larger importance for the MS class, while the third is protective, with higher values contributing negatively to the MS class probability.

\section{Conclusion}
In this study we developed an ML pipeline to analyze microarray data from MS patients and Control, showing how ML combined with xAI can compare and enhance results from canonical DEA. Our results show that xAI techniques like SHAP not only can confirm findings from DEA but also may detect additional genes which could play important roles in MS pathogenesis. These insights have been validated through an enrichment analysis, showing how genes important for SHAP are involved in pathways like sphingolipid signaling, Th1/Th2/Th17 cell differentiation, and Epstein-Barr virus infection, all  biologically relevant to MS.

\section*{\bf Conflict of interests}
\label{sec:CONFLICT-OF-INTERESTS}
The authors declare no conflict of interest.

\section*{\bf Acknowledgments}
We gratefully acknowledge the support of the CINI InfoLife laboratory in this research

\section*{\bf Funding}
\label{sec:FUNDING}
This work was supported by the co-funding European Union - Next Generation
EU, in the context of The National Recovery and Resilience Plan, Mission 4 Component 2, Investment 1.1, Call PRIN 2022 D.D. 104 02-02-2022 – MEDICA
Project, CUP N. I53D23003720 006 and Investment 1.5 Ecosystems of Innovation, Project Tuscany Health Ecosystem (THE), CUP: B83C22003920001, Spoke 3, by the SPARK programme at the University of Pisa, by the project "Hub multidisciplinare e interregionale di ricerca e sperimentazione clinica per il contrasto alle pandemie e all’antibioticoresistenza (PAN-HUB)” funded by the Italian Ministry of Health (POS 2014-2020, project ID: T4-AN-07, CUP: I53C22001300001).
S. G. Galfrè is supported in part by a 2023 NARSAD Young Investigator Grant from the Brain \& Behavior Research Foundation.

\section*{\bf Availability of data and software code}
\label{sec:AVAILABILITY}
Code and data available at: 
 \indent \url{https://github.com/seriph78/ML_for_MS.git}

\footnotesize
\bibliographystyle{unsrt}
\bibliography{Bibliography.bib} 
\normalsize

\end{document}